%% file: main.tex
\documentclass[runningheads]{llncs}

\usepackage[T1]{fontenc}
\usepackage{graphicx}
\usepackage{booktabs}
\usepackage{multirow}
\usepackage{amsmath,amssymb}
\usepackage{xcolor}
\usepackage[
  hidelinks,
  pdftitle={Identify, Locate, Link: End-to-End Key-Value Extraction from Document Images},
  pdfauthor={A. Said Gurbuz, Ahmed Nassar, Christoph Auer, Maksym Lysak, Lucas Morin, Matteo Omenetti, Tim Strohmeyer, Panagiotis Vagenas, Nikolaos Livathinos, Michele Dolfi, and Peter Staar},
  pdfsubject={End-to-end key-value extraction from document images},
  pdfkeywords={key-value extraction, document understanding, vision-language models, layout-aware evaluation}
]{hyperref}
\usepackage{cleveref}
\usepackage{subcaption}
\usepackage[most]{tcolorbox}
\usepackage{tikz}
\usetikzlibrary{positioning,arrows.meta}
\usepackage{placeins}
\usepackage{colortbl}

\definecolor{hlrow}{RGB}{232,245,233}

\definecolor{structtag}{RGB}{63,81,181}
\definecolor{contenttag}{RGB}{76,175,80}
\definecolor{loctag}{RGB}{255,152,0}
\definecolor{linktag}{RGB}{233,30,99}
\definecolor{texttag}{RGB}{97,97,97}

\newcommand{\tagtt}[1]{\texttt{#1}}

\newcommand{\cmark}{\textcolor{green!60!black}{\checkmark}}
\newcommand{\pmark}{\textcolor{orange!85!black}{\checkmark\textsuperscript{\scriptsize *}}}
\newcommand{\xmark}{\textcolor{red}{$\times$}}

\begin{document}

\title{Identify, Locate, Link: End-to-End Key-Value Extraction from Document Images}
\titlerunning{Identify, Locate, Link: End-to-End Key-Value Extraction}

\author{A. Said Gurbuz\inst{1,2} \and
Ahmed Nassar\inst{1} \and
Christoph Auer\inst{1} \and
Maksym Lysak\inst{1} \and
Lucas Morin\inst{1} \and
Matteo Omenetti\inst{1} \and
Tim Strohmeyer\inst{1} \and
Panagiotis Vagenas\inst{1} \and
Nikolaos Livathinos\inst{1} \and
Michele Dolfi\inst{1} \and
Peter Staar\inst{1}}
\authorrunning{A. Said Gurbuz et al.}
\institute{IBM Research Zurich, R\"uschlikon, Switzerland \and
ETH Zurich, Zurich, Switzerland\\
\email{said.guerbuez@inf.ethz.ch}}

\maketitle

\input{sections/abstract}

\input{sections/introduction}

\input{sections/related_work}

\input{sections/method}

\input{sections/experiments}

\input{sections/conclusion}

\FloatBarrier
\bibliographystyle{splncs04}
\bibliography{references}

\end{document}

%% file: sections/abstract.tex
\begin{abstract}
Document processing pipelines traditionally cascade optical character recognition (OCR) engines with downstream models for structured information extraction, leading to multi-stage error propagation. We fine-tune SmolDocling, a compact 256M-parameter vision-language model (VLM), to perform end-to-end key-value extraction directly from document images, jointly solving identification, localization, and association in a single pass without OCR preprocessing. We extend DocTags with specialized key, value, region, and link tags, enabling many-to-many relationships in a unified output sequence. To address data limitations, we design an augmentation pipeline combining synthetic form filling and graph-based crops that preserve complete key-value subgraphs. We further introduce a layout-aware evaluation framework extending text matching with spatial bounding box verification. On FUNSD, XFUND, and a large-scale private dataset, our model outperforms larger zero-shot VLM baselines under layout-aware evaluation, while being 27$\times$ smaller than Qwen2.5-VL~(7B) and over 5$\times$ faster at inference. The model weights will be released publicly after publication.

\keywords{Key-value extraction \and Document understanding \and Vision-language models \and End-to-end \and Layout-aware evaluation}
\end{abstract}

%% file: sections/introduction.tex
\section{Introduction}
\label{sec:introduction}

\begin{figure}[t]
  \centering
  \begin{subfigure}[b]{0.48\textwidth}
    \centering
    \tcbox[colframe=black, arc=6pt, boxrule=1pt, left=0pt, right=0pt, top=0pt, bottom=0pt]{%
      \includegraphics[height=0.27\textheight, keepaspectratio]{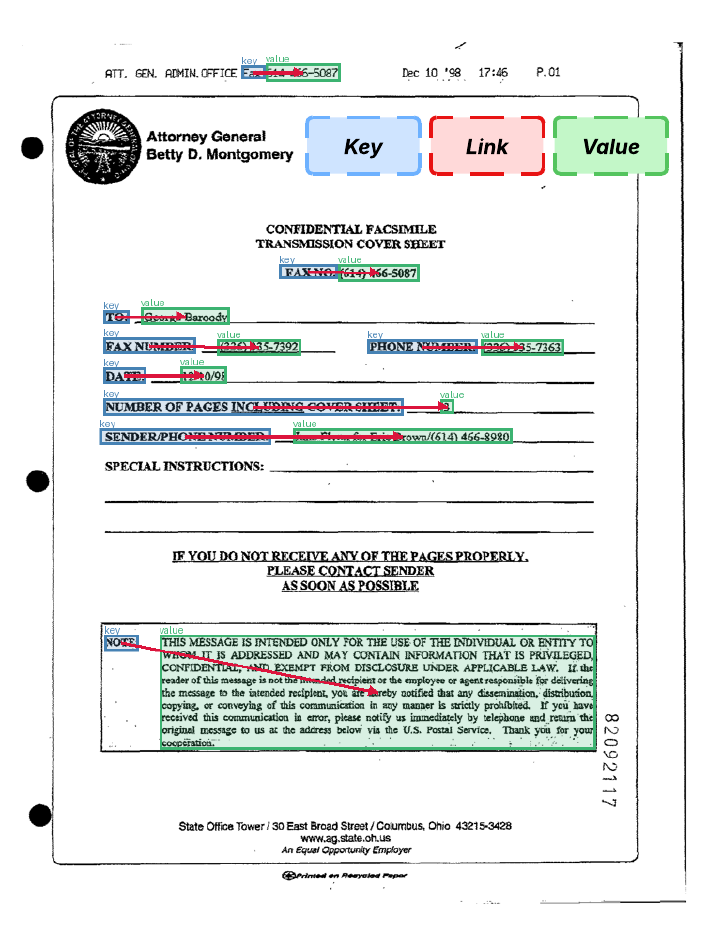}%
    }
    \caption{Key-value extraction with links}
    \label{fig:kv-highlighted}
  \end{subfigure}
  \hfill
  \begin{subfigure}[b]{0.48\textwidth}
    \centering
    \tcbox[colframe=black, arc=6pt, boxrule=1pt, left=0pt, right=0pt, top=0pt, bottom=0pt]{%
      \includegraphics[height=0.27\textheight, keepaspectratio]{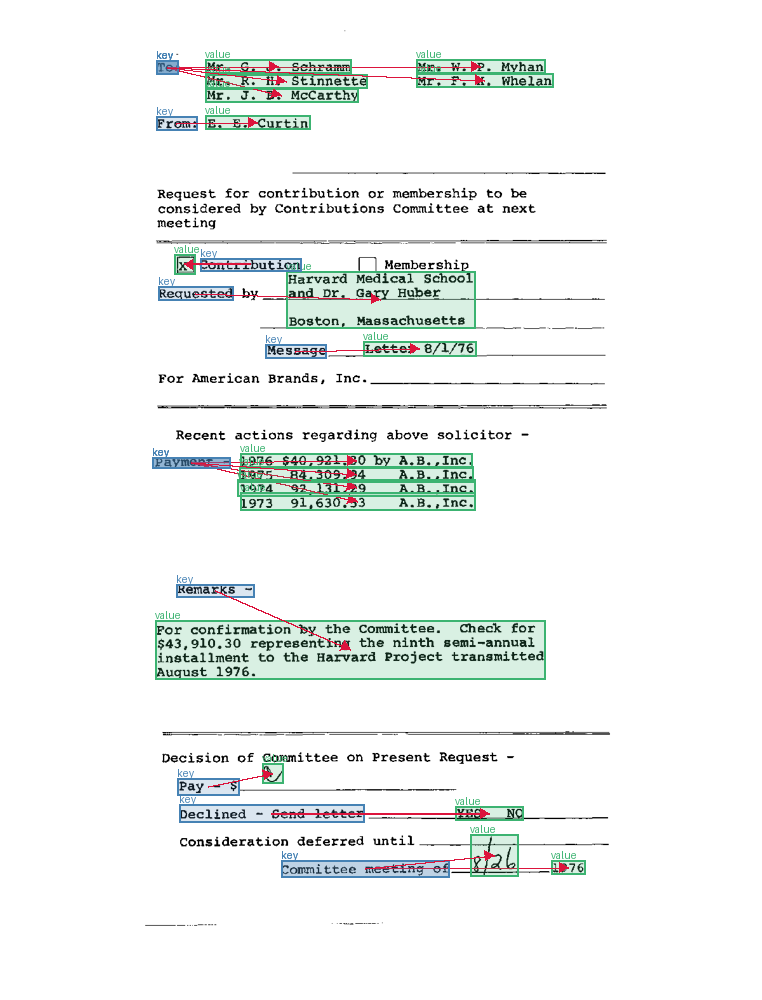}%
    }
    \caption{Complex many-to-many relationships}
    \label{fig:kv-complex}
  \end{subfigure}
  \caption{Key-value extraction examples: (a)~visualization with keys (blue), values (green), and links (red arrows); (b)~complex many-to-many relationships forming a directed graph.}
  \label{fig:kv-example}
\end{figure}

Over 80\% of enterprise information resides in unstructured documents~\cite{king2019idcdata}, ranging from invoices and contracts to medical forms and regulatory filings, yet structured knowledge extraction from these documents remains a major bottleneck.
Unlike plain optical character recognition~(OCR), which recovers text alone, key-value extraction requires jointly understanding textual content, spatial layout, and semantic relationships.
\Cref{fig:kv-highlighted} illustrates a concrete example: given a scanned form, the system must identify which text regions are keys and which are values, localize each with a bounding box, and establish the correct links between them.

The task encompasses three interconnected subtasks: \emph{identification} of text elements as keys or values (semantic entity recognition), \emph{localization} of their spatial positions via bounding boxes (grounding), and \emph{association} of related key-value pairs (relation extraction).
Real documents further complicate matters with many-to-many relationships where a single key maps to multiple values, or the values themselves serve as keys to sub-options, forming directed graph structures (\cref{fig:kv-complex}).


\paragraph{Limitations of current approaches.}
State-of-the-art systems~\cite{huang2022layoutlmv3,xu2021layoutxlm,luo2023geolayoutlm} employ multi-stage pipelines that first extract text and layout through an external OCR engine~\cite{smith2007tesseract}, then apply encoder models for token classification and relation extraction (RE).
While effective, this cascade architecture introduces error propagation, language dependence from OCR availability~\cite{xu2021layoutxlm}, engineering complexity from coordinating multiple components, and fragmented outputs that require post-processing.

\paragraph{Motivation.}
We address these limitations with an end-to-end approach that directly processes document images without requiring any external OCR or layout analysis.
Our approach adopts and extends the DocTags format~\cite{nassarsmoldocling2025}, a structured markup language for document understanding, with four specialized tags for key-value pairs and explicit linking mechanisms.
This unified representation eliminates task-specific post-processing while maintaining the flexibility to express complex many-to-many relationships through unique identifiers and directed links.
Because localization grounds predictions to source regions, the output supports repeated-field disambiguation and downstream visual retrieval workflows.

\paragraph{Contributions.}
This paper makes four contributions:
\begin{enumerate}
  \item \textbf{End-to-end key-value extraction.}
  We fine-tune a 256M VLM to perform identification, localization, and association of key-value pairs in a single forward pass without OCR or layout analysis, showing that 
  our model is competitive with larger OCR-free foundation models such as Qwen2.5-VL~\cite{bai2025qwen25vl}~(7B) while being $27\times$ smaller.
  \item \textbf{DocTags extension for key-value semantics.}
  We extend the DocTags vocabulary with four tags for key-value regions, keys, values, and directed links, with unique identifiers supporting many-to-many links in a unified output format.
  \item \textbf{Data augmentation pipeline.}
  We combine synthetic form filling with graph-based crops that preserve complete key-value subgraphs, improving relation extraction F1 by up to 41\%.
  \item \textbf{Layout-aware evaluation framework.}
  We introduce metrics combining text matching with spatial bounding box verification via Intersection over Union (IoU $\geq$ 0.7), providing more realistic assessment than text-only metrics that ignore spatial correctness.
\end{enumerate}

Experiments across FUNSD~\cite{jaumefunsd2019}, XFUND~\cite{xu-2022-xfund}, and a large-scale private dataset show that our 256M model outperforms larger zero-shot VLM baselines under layout-aware evaluation, while being 27$\times$ smaller than Qwen2.5-VL~(7B) and over 5$\times$ faster at inference.

%% file: sections/related_work.tex
\section{Related Work}
\label{sec:related-work}

Key-value extraction from documents has been approached through four main paradigms, each with different trade-offs between accuracy, complexity, and OCR dependence.
\Cref{tab:related-work-comparison} summarizes representative models across these paradigms.

\begin{table}[t]
  \caption{Overview of key-value extraction models. OCR-free = no external OCR needed; Loc.\ = predicts bounding boxes; Struct.\ = structured output format.}
  \label{tab:related-work-comparison}
  \centering
  \small
  \setlength{\tabcolsep}{3.5pt}
  \begin{tabular}{llcccc}
    \toprule
    \textbf{Paradigm} & \textbf{Model} & \textbf{OCR-free} & \textbf{Loc.} & \textbf{Struct.} & \textbf{Params} \\
    \midrule
    \multirow{3}{*}{Multimodal enc.}
      & LayoutLMv3~\cite{huang2022layoutlmv3}  & \xmark & \xmark & \xmark & 133M/368M \\
      & LayoutXLM~\cite{xu2021layoutxlm}        & \xmark & \xmark & \xmark & 345M/625M \\
      & LiLT~\cite{wang2022lilt}                & \xmark & \xmark & \xmark & 131M \\
    \midrule
    \multirow{3}{*}{Structure-aware}
      & FormNet~\cite{lee2022formnet}            & \xmark & \xmark & \xmark & 217M  \\
      & GeoLayoutLM~\cite{luo2023geolayoutlm}   & \xmark & \xmark & \xmark & 399M \\
      & XFormParser~\cite{cheng2024xformparser}  & \xmark & \xmark & \xmark & 370M \\
    \midrule
    \multirow{2}{*}{Decoder-only}
      & DocLLM~\cite{wang2023docllm}             & \xmark & \xmark & \xmark & 7B   \\
      & LayTextLLM~\cite{lu2025laytextllm}       & \xmark & \cmark & \xmark & 7B   \\
    \midrule
    End-to-end
      & Donut~\cite{kim2022donut}                & \cmark & \xmark & \pmark & 200M \\
    \rowcolor{hlrow}
    End-to-end
      & \textbf{Ours (SmolDocling)}                      & \cmark & \cmark & \cmark & 256M \\
    \bottomrule
  \end{tabular}
  \\[2pt]
  {\scriptsize $^*$ structured output, but no complex key-value relationship support.}
\end{table}

\paragraph{OCR-based multimodal transformers.}
The dominant paradigm combines OCR-extracted text with visual and layout features through multimodal pre-training, building on language encoders~\cite{devlin2019bert,liu2019roberta,conneau2020xlmroberta,chi2021infoxlm}.
The LayoutLM series~\cite{Xu_2020layoutlm,xu2022layoutlmv2,huang2022layoutlmv3} progressively unified these modalities, culminating in LayoutLMv3's joint text-image masking with visual patch embeddings~\cite{huang2022layoutlmv3}.
Cross-lingual variants such as LayoutXLM~\cite{xu2021layoutxlm} extend this to multilingual documents by pre-training across seven languages, while language-independent approaches~\cite{wang2022lilt,hong2022bros,li2021structext} decouple layout from language modeling to enable cross-lingual transfer without multilingual pre-training.
These models achieve strong results on form understanding benchmarks~\cite{jaumefunsd2019,xu-2022-xfund} by adding relation extraction heads on top of entity recognition.
However, they frame key-value extraction as token classification over pre-extracted text, meaning that they cannot recover from upstream OCR errors and are constrained to languages with reliable OCR support.

\paragraph{Structure-aware models.}
A complementary line of work enriches transformers with explicit spatial inductive biases, motivated by the observation that standard attention mechanisms learn geometric relationships only implicitly~\cite{luo2023geolayoutlm}.
Graph-based approaches construct adjacency graphs over OCR tokens and refine them through graph convolutional networks~\cite{lee2022formnet,Cao_2023_graphRevisedIE}, while geometry-focused pre-training introduces objectives that directly model spatial relationships between entities~\cite{luo2023geolayoutlm}.
Joint entity-relation architectures further improve linking accuracy by combining sequence labeling with relation prediction in a single framework~\cite{cheng2024xformparser}.
These models confirm that spatial structure is critical for key-value linking, yet they inherit the same OCR dependency as the multimodal encoder paradigm.

\paragraph{Decoder-only models.}
With the rise of large language models, recent work has explored injecting layout awareness into decoder-only architectures.
Approaches range from disentangled spatial attention over OCR text~\cite{wang2023docllm} to interleaving projected bounding box tokens directly into the input sequence~\cite{lu2025laytextllm}.
While these models leverage powerful generative capabilities and scale to billions of parameters, they still require pre-extracted text and produce output in natural language or token-tag formats that need post-processing to yield structured key-value pairs.

\paragraph{End-to-end approaches.}
Donut~\cite{kim2022donut} demonstrated the feasibility of OCR-free document understanding with a vision transformer encoder and autoregressive decoder. While it can serialize outputs in JSON-like form, it does not provide spatial grounding and does not explicitly model complex key-value relationship structures such as one-to-many or many-to-many link graphs.
Nougat~\cite{blecher2023nougat} extended the encoder-decoder paradigm for academic document conversion, though it targets document-level Markdown generation rather than structured information extraction.
TrOCR~\cite{li2022trocr} and Docling~\cite{auer2024docling} further extend encoder-decoder architectures to richer OCR and document conversion tasks, but focus on text recognition and serialization rather than structured key-value extraction.
SmolDocling~\cite{nassarsmoldocling2025} introduced the DocTags format, a unified markup language that explicitly represents both structure and spatial information through normalized bounding box tokens, along with an ultra-compact 256M model pre-trained on web-scale document images.
However, SmolDocling was designed for general document conversion (tables, text blocks, captions) and lacks tags for key-value relationships.
This work extends SmolDocling to key-value extraction by introducing four specialized tags for key-value regions, keys, values, and directed links. We further propose a data augmentation pipeline that addresses training data scarcity through synthetic form filling and graph-based crop generation, and a layout-aware evaluation framework that verifies spatial correctness alongside text matching.

%% file: sections/method.tex
\section{Method}
\label{sec:method}

\subsection{Problem Definition}
\label{sec:problem-definition}

Given a document image $\mathcal{I} \in \mathbb{R}^{H \times W \times 3}$, key-value extraction requires jointly solving three subtasks:
(1)~\emph{Identification}: classifying textual elements into keys or values, analogous to named entity recognition but operating directly on visual input;
(2)~\emph{Localization}: predicting bounding boxes $\mathcal{B}_i = (x_1, y_1, x_2, y_2)$ for each identified element;
and (3)~\emph{Association}: establishing directed relationships between keys and values, where many-to-many mappings are permitted.
The complete extraction objective combines all three:
\begin{equation}
  \mathcal{F}: \mathcal{I} \rightarrow \{(k_j, v_k, \mathcal{B}_j, \mathcal{B}_k) : (k_j, v_k) \in \mathcal{R}\}
  \label{eq:joint-objective}
\end{equation}
where $(k_j, v_k)$ denotes a key-value relationship, $\mathcal{B}_j$ and $\mathcal{B}_k$ are their respective bounding boxes, and $\mathcal{R}$ supports many-to-many links, forming a directed graph $G = (\mathcal{K} \cup \mathcal{V}, \mathcal{R})$.

\subsection{DocTags Representation}
\label{sec:doctags}

We adopt the DocTags markup format~\cite{nassarsmoldocling2025} for structured document conversion.
DocTags provide a standardized vocabulary of XML-style tags that explicitly separate textual content from document structure~\cite{nassarsmoldocling2025,lysak2023otsl}, reducing ambiguity compared to direct conversion to HTML or Markdown.
Each document element incorporates location information through four consecutive bounding box tokens \texttt{<loc\_$x_1$><loc\_$y_1$><loc\_$x_2$><loc\_$y_2$>}, where coordinates are normalized to $[0, 500]$ for scale invariance.

\paragraph{Key-value extension.}
We extend the DocTags vocabulary with four specialized tags for key-value extraction:
\begin{itemize}
  \item \textbf{\tagtt{<key\_value\_region>}}: Encapsulates a region containing related key-value pairs and its bounding box.
  \item \textbf{\tagtt{<key\_$i$>}}: Marks a key element with unique identifier $i \in \mathbb{N}$.
  \item \textbf{\tagtt{<value\_$j$>}}: Marks a value element with unique identifier $j \in \mathbb{N}$.
  \item \textbf{\tagtt{<link\_$k$>}}: Establishes a directed link from the enclosing element to the element with identifier~$k$.
\end{itemize}

\begin{figure}[t]
  \centering
  \begin{tikzpicture}[node distance=3cm]
    \node[inner sep=1, draw=black, thick, rounded corners=2pt] (image) {\includegraphics[width=0.30\textwidth]{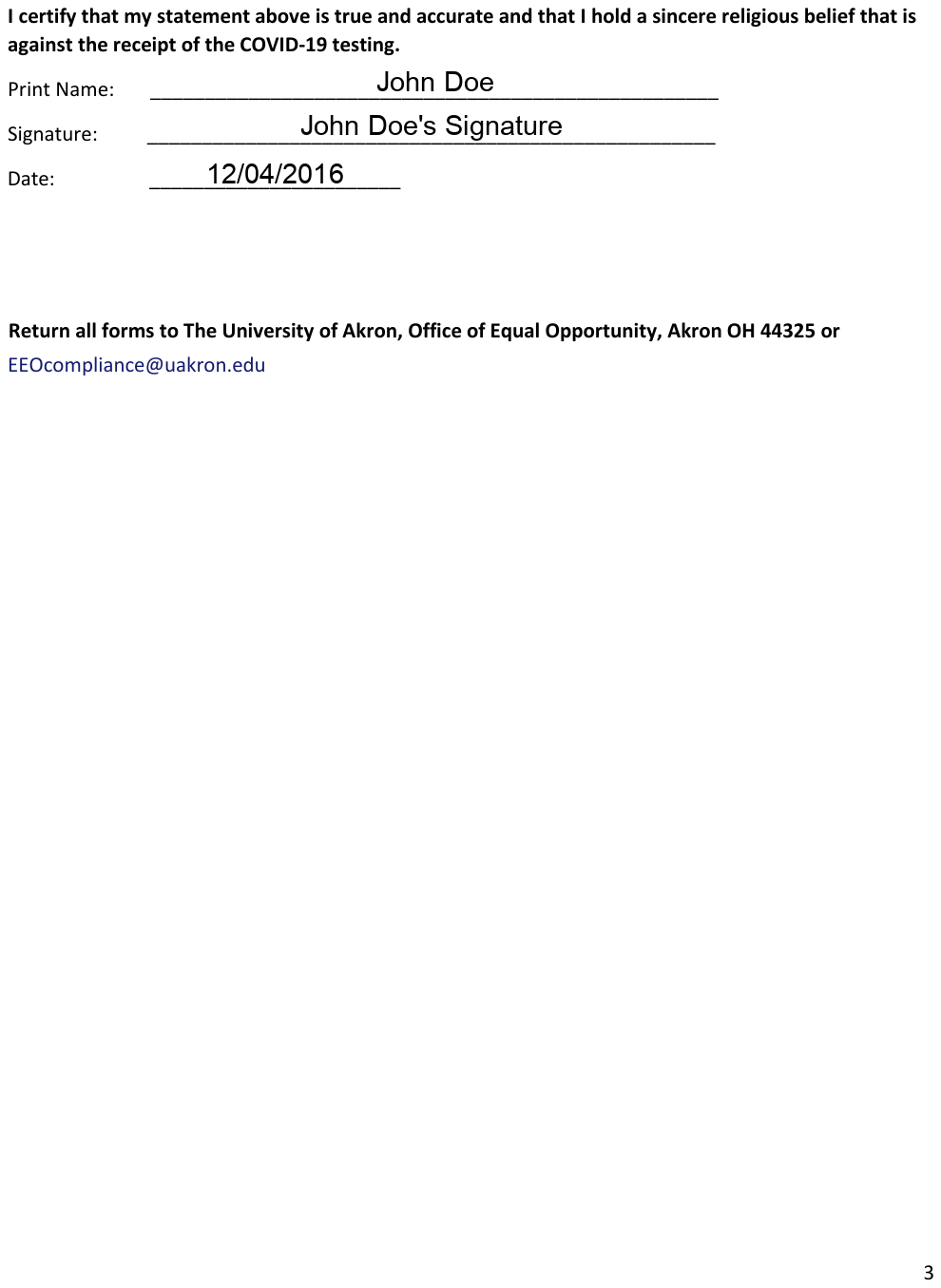}};
    \node[inner sep=0, right=2cm of image] (doctag) {\includegraphics[width=0.52\textwidth]{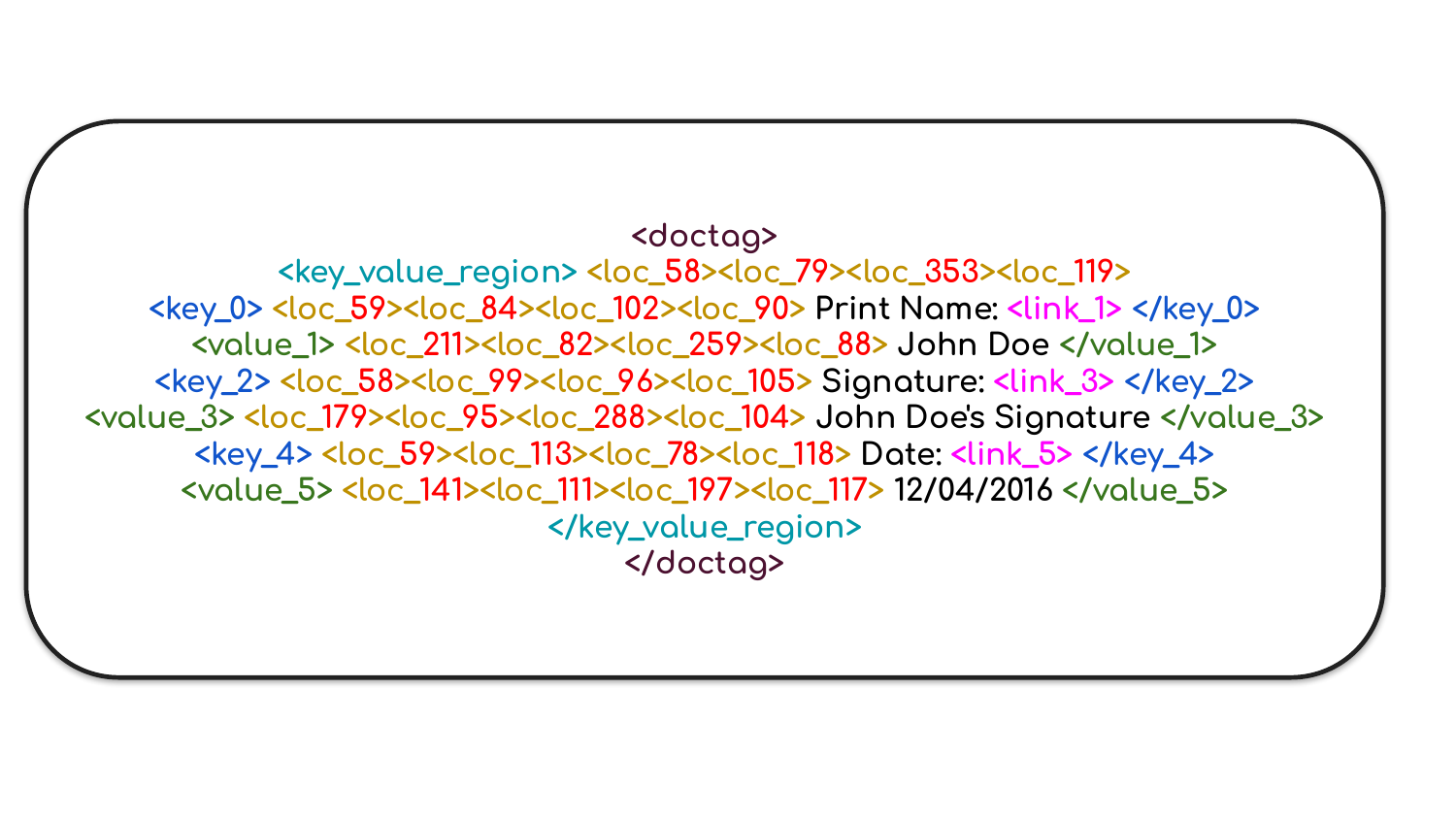}};
    \draw[-Stealth, line width=1.2pt] (image.east) -- node[above, font=\small\bfseries, yshift=2pt] {DocTags} (doctag.west);
  \end{tikzpicture}
  \caption{DocTags representation: a document image (left) and its corresponding structured markup (right) encoding key-value content, bounding boxes, and links.}
  \label{fig:doctag-overview}
\end{figure}

\noindent
Each key and value element appears exactly once with a unique identifier; many-to-many relationships are expressed by placing multiple \tagtt{<link>} tokens inside a single key.
By default, link tokens are placed within key elements (\emph{link-in-key} strategy), anchoring the link before the target value is decoded.
We also explore an alternative \emph{link-in-value} strategy (links in values) and spatial ordering of elements by top-left coordinates~\cite{wang2021layoutreader} in our ablation studies (\cref{sec:ablations}).
At inference time, the parser only converts well-formed \tagtt{<key\_$i$>} and \tagtt{<value\_$i$>} spans into graph cells; links use document-local identifiers, and references to missing target identifiers are discarded.
Duplicate or cyclic links are not repaired, and therefore remain extra predicted edges during evaluation unless supported by the ground truth.
\Cref{fig:doctag-overview} illustrates the complete DocTags representation for a document with key-value pairs.

\subsection{Model Architecture}
\label{sec:architecture}

We build on SmolDocling~\cite{nassarsmoldocling2025}, an ultra-compact vision language model pre-trained on web-scale document images for end-to-end document conversion.
The model follows the SmolVLM~\cite{marafiotismolvlm2025} architecture and comprises two components.

The \textbf{vision encoder} is a SigLIP~\cite{zhaisiglip2023} model with patch size 16 (93M parameters).
It splits the input image into non-overlapping patches (at 384$\times$384 resolution), processes them through a vision transformer, and produces patch embeddings $\mathbf{H}^{\mathrm{vis}} \in \mathbb{R}^{P \times d}$.
A learned linear projection maps these embeddings into the decoder's token space, followed by a pixel-shuffle pooling operation that reduces the sequence length by a factor of 4, yielding compact visual tokens.

The \textbf{language decoder} is SmolLM2~\cite{allal2025smollm2} (135M parameters), a decoder-only transformer~\cite{vaswani2023attentionisallyouneed} that autoregressively generates the DocTags token sequence conditioned on the visual tokens.
The vocabulary is extended with special tokens for key-value tags, location tokens ($\texttt{<loc\_0>}$ through $\texttt{<loc\_500>}$), and link identifiers.

\begin{figure}[t]
  \centering
  \includegraphics[trim={4.5cm 3.75cm 4.5cm 3.75cm}, clip, width=0.88\textwidth]{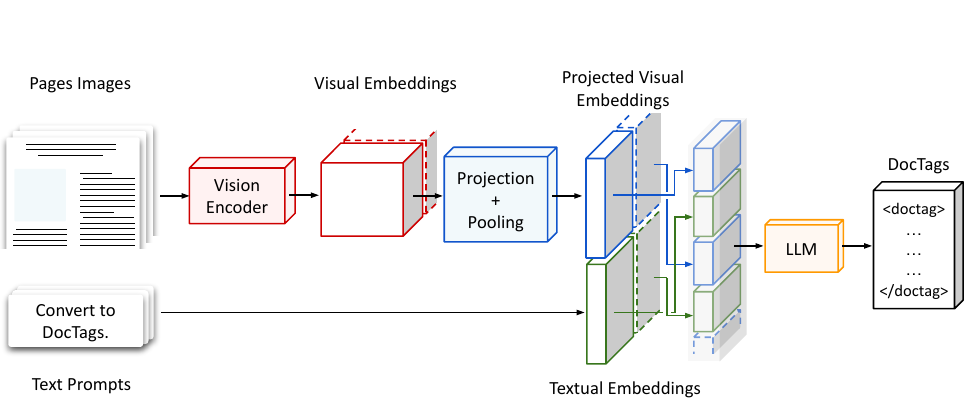}
  \caption{Model architecture \cite{nassarsmoldocling2025}: a SigLIP vision encoder (93M) produces patch embeddings that are projected and combined with text embeddings, then fed to a SmolLM2 decoder (135M) to generate DocTags output.}
  \label{fig:architecture}
\end{figure}

The total model size is 256M parameters, making it suitable for on-device deployment~\cite{xu2024ondevice}.
During fine-tuning, all parameters (encoder, projector, and decoder) are unfrozen.
\Cref{fig:architecture} shows the architecture overview.

\subsection{Training Objective}
\label{sec:training}

We train with standard cross-entropy loss over the DocTags token sequence, conditioned on visual tokens.

\paragraph{Token-aware weighted loss.}
We also explore a variant that assigns structural tokens (e.g., \tagtt{<key\_value\_region>}, \tagtt{<key>}, \tagtt{<value>}, \tagtt{<link>}) a weight of $\lambda = 5$.
This up-weighting prevents frequent content tokens from dominating the loss and encourages accurate generation of the key-value graph structure; its effect is evaluated in our ablation studies (\cref{sec:ablations}).

\subsection{Datasets}
\label{sec:datasets}

We use three datasets that satisfy the strict requirements for key-value extraction: element-level bounding box annotations, semantic key/value labels, and explicit linking information between entities.
\Cref{tab:datasets} summarizes dataset statistics.

\textbf{FUNSD}~\cite{jaumefunsd2019} contains 199 noisy scanned forms with word-level annotations across four semantic classes (question, answer, header, other) and explicit key-value linking.
The dataset is small but challenging due to diverse layouts and significant scanning noise.

\textbf{XFUND}~\cite{xu-2022-xfund} extends the FUNSD annotation scheme to seven languages (Chinese, Japanese, Spanish, French, Italian, German, Portuguese) with 1,393 human-annotated forms, providing a multilingual benchmark for cross-lingual key-value extraction.

\textbf{DocLayNetV2}~\cite{Pfitzmann_2022_doclaynet} is a large-scale private dataset with 20,346 pages containing key-value pairs, filtered from a 100K-page human-annotated collection spanning 20 languages and diverse document types.
Pages are sampled from an internal 1.4M-page crawl to maximize layout and topic diversity.

\begin{table}[t]
  \caption{Dataset statistics. Entities and Links columns report totals for all splits.}
  \label{tab:datasets}
  \centering
  \setlength{\tabcolsep}{6pt}
  \renewcommand{\arraystretch}{1.15}
  \begin{tabular}{l rrr rr}
    \toprule
    \multirow{2}{*}{\textbf{Dataset}} & \multicolumn{3}{c}{\textbf{Samples}} & \multicolumn{2}{c}{\textbf{Annotations}} \\
    \cmidrule(lr){2-4} \cmidrule(lr){5-6}
    & \textbf{Train} & \textbf{Val} & \textbf{Test} & \textbf{Entities} & \textbf{Links} \\
    \midrule
    DocLayNetV2       & 16,229         & 2,032        & 2,085         & 429,595           & 241,469        \\
    XFUND             & 1,043          & 175          & 175           & 69,974            & 51,146         \\
    FUNSD             & 149            & 25           & 25            & 8,362             & 5,621          \\
    \bottomrule
  \end{tabular}
\end{table}

\begin{figure}[t]
  \centering

  \begin{subfigure}[t]{\textwidth}
    \centering
    \begin{tikzpicture}
      \node[draw=black!70, dashed, line width=1.2pt, rounded corners=1pt, inner sep=0.9pt, minimum width=0.92\linewidth, align=center]
      {\includegraphics[width=0.86\linewidth, trim=110pt 28pt 110pt 122pt, clip]{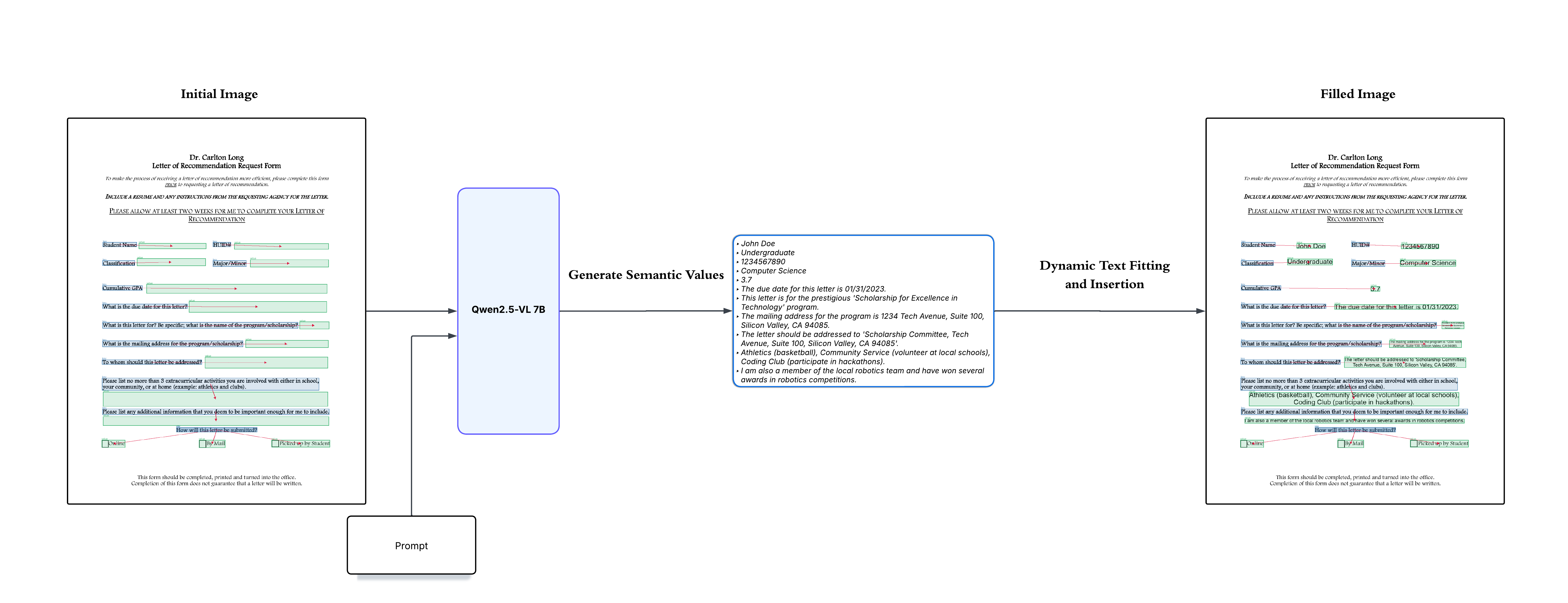}};
    \end{tikzpicture}
    \caption{Synthetic form filling.}
    \label{fig:form-filling}
  \end{subfigure}

  \vspace{0em}

  \begin{subfigure}[t]{\textwidth}
    \centering
    \begin{tikzpicture}
      \node[draw=black!70, dashed, line width=1.2pt, rounded corners=1pt, inner sep=0.9pt, minimum width=0.92\linewidth, align=center]
      {\includegraphics[width=0.55\linewidth]{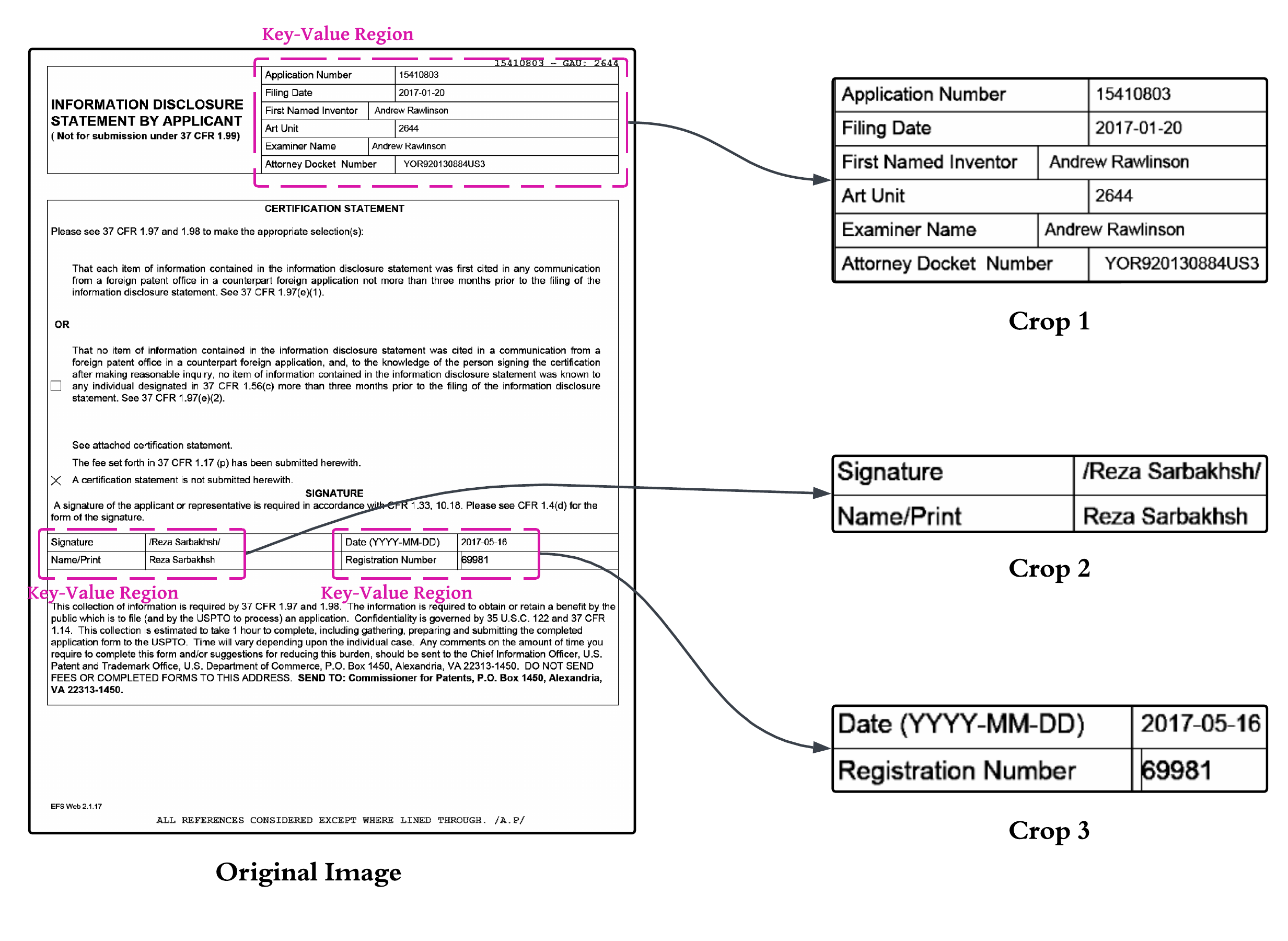}};
    \end{tikzpicture}
    \caption{Crop augmentation.}
    \label{fig:crop-generation}
  \end{subfigure}
  \caption{Data augmentation pipeline used for training: (a) synthetic filling of empty form fields and (b) graph-based crop generation preserving complete key-value subgraphs.}
  \label{fig:data-pipeline-overview}
\end{figure}

\paragraph{Synthetic form filling.}
DocLayNetV2 contains empty form fields (unfilled areas) that cause the model to learn predicting empty values, reducing its ability to generalize.
We address this by providing the full document image and associated key texts to Qwen2.5-VL~72B, which generates contextually appropriate values.
The generated text is rendered into the empty regions using adaptive font sizing with multi-line wrapping to respect spatial constraints, and bounding box annotations are updated accordingly.
\Cref{fig:form-filling} illustrates this process.

\paragraph{Crop augmentation.}
Full document pages contain many elements beyond key-value pairs.
To help the model focus on key-value semantics, we generate focused crops by:
(i)~building a graph of key-value pairs connected by their link relationships;
(ii)~computing connected components and merging spatially proximate ones within a distance threshold;
(iii)~adding random padding around each component while respecting image boundaries.
This produces training samples that contain complete key-value subgraphs without fragmenting semantic units across crops.
\Cref{fig:crop-generation} illustrates this process.

\paragraph{Image transformations.}
During training, we apply on-the-fly augmentations using Albumentations~\cite{buslaev2020albumentations}: erosion and dilation (adapted from Nougat~\cite{blecher2023nougat}), Gaussian noise, Gaussian blur, color jitter, bitmap thresholding, and JPEG compression artifacts.
All transformations preserve bounding box validity.

\subsection{Evaluation Framework}
\label{sec:evaluation}

Traditional evaluation for key-value extraction~\cite{xu2021layoutxlm,huang2022layoutlmv3} covers two subtasks: semantic entity recognition (SER), which compares predicted and ground-truth entities as (text, label) tuples requiring matching text content and semantic label (key or value) with a greedy one-to-one assignment to prevent double-counting; and relation extraction (RE), where both the source and target entities of a relation must match.
This text-only protocol is appropriate for encoder-based models that perform token classification over pre-extracted OCR text, since spatial positions are given as input and need not be verified in the output.

However, text-only evaluation is insufficient for end-to-end models that must simultaneously perform OCR, classification, and localization.
Consider a model that correctly predicts the text ``Name'' with the label \emph{key}, but localizes it 200 pixels away from the actual field; text-only evaluation counts this as correct, despite the prediction being spatially meaningless.
To address this, we introduce a layout-aware evaluation framework that extends text matching with spatial verification.

\paragraph{Layout-aware matching.}
An entity is considered a true positive only when three conditions are jointly satisfied: (i)~the predicted text matches the ground truth (using flexible matching with Levenshtein similarity~\cite{levenshtein_1966} $\rho_{\text{Lev}} \geq 0.8$), (ii)~the semantic labels agree, and (iii)~the spatial overlap exceeds a threshold:
\begin{equation}
  \text{IoU}(\mathcal{B}_p, \mathcal{B}_g) = \frac{\text{Area}(\mathcal{B}_p \cap \mathcal{B}_g)}{\text{Area}(\mathcal{B}_p \cup \mathcal{B}_g)} \geq 0.7
\end{equation}
For relations, both the source key and target value must satisfy all three conditions.
To ensure globally optimal entity assignment (important when the same text appears at multiple locations), we use the Hungarian algorithm~\cite{kuhn1955hungarian} for bipartite matching, minimizing negative IoU over compatible text-label pairs.

\paragraph{Protocol selection.}
In our experiments, layout-aware evaluation is used for all models that predict bounding boxes (our model and Qwen2.5-VL), while text-only evaluation is used for models that output text without spatial information (GPT-4o, Llama-3.2).
This distinction is noted in all results tables.

%% file: sections/experiments.tex
\section{Experiments}
\label{sec:experiments}

\subsection{Setup}
\label{sec:setup}

All experiments were conducted on 8 NVIDIA A100 GPUs by fine-tuning the pre-trained SmolDocling-256M checkpoint~\cite{nassarsmoldocling2025} for 10 epochs; the best checkpoint was selected by validation loss.
Key hyperparameters: effective batch size 32 (per-device 2, gradient accumulation 2), AdamW optimizer with cosine scheduling, 3\% warmup, gradient clipping at 1.0.
We used different learning rates: $2{\times}10^{-4}$ for the decoder, $2{\times}10^{-6}$ for the vision encoder, and $7{\times}10^{-6}$ for the vision projector.
All model parameters were unfrozen during training.

\subsection{Baselines}
\label{sec:baselines}

We report contextual OCR/text-layout references and direct raw-image VLM comparisons.
OCR/text-layout models consume pre-extracted text and boxes, so they are upper-bound references rather than direct layout-aware comparisons.
For raw-image comparison, we evaluate Llama-3.2-11B~\cite{grattafiori2024llama3}, GPT-4o~\cite{openai2024gpt4}, and Qwen2.5-VL-7B~\cite{bai2025qwen25vl} zero-shot with task-specific prompts.
Among these, only Qwen2.5-VL supports bounding box prediction, enabling layout-aware evaluation alongside our model.

\paragraph{Evaluation protocol.}
For SmolDocling and Qwen, layout-aware metrics require text, label, and spatial agreement (IoU $\geq$ 0.7).
Models without localization (Llama-3.2, GPT-4o, Donut-style outputs) use text-only evaluation.
OCR/text-layout models cannot use the same layout-aware metric because spatial positions are inputs, not predictions.

\subsection{Results}
\label{sec:results}

\begin{table}[t]
  \caption{Contextual OCR/text-layout baselines on FUNSD and XFUND. These models consume OCR text and/or layout as input and are not directly comparable under our layout-aware metric.}
  \label{tab:ocr-baselines}
  \centering
  \scriptsize
  \setlength{\tabcolsep}{3pt}
  \renewcommand{\arraystretch}{0.95}
  \begin{tabular}{lccccc}
    \toprule
    \multirow{2}{*}{\textbf{Model}} & \multirow{2}{*}{\textbf{Input}} & \multicolumn{2}{c}{\textbf{FUNSD}} & \multicolumn{2}{c}{\textbf{XFUND}} \\
    & & \textbf{SER} & \textbf{RE} & \textbf{SER} & \textbf{RE} \\
    \midrule
    BERT$_\text{base}$~\cite{devlin2019bert}               & T     & 0.610 & 0.277 & --    & --    \\
    XLM-R$_\text{base}$~\cite{conneau2020xlmroberta}       & T     & 0.667 & 0.266 & 0.342 & 0.222 \\
    XLM-R$_\text{large}$~\cite{conneau2020xlmroberta}      & T     & 0.707 & 0.347 & 0.429 & 0.266 \\
    InfoXLM$_\text{base}$~\cite{chi2021infoxlm}            & T     & 0.685 & 0.292 & 0.373 & 0.235 \\
    InfoXLM$_\text{large}$~\cite{chi2021infoxlm}           & T     & 0.733 & 0.368 & 0.448 & 0.302 \\
    LayoutLMv2$_\text{base}$~\cite{xu2022layoutlmv2}       & I+T+L & 0.819 & 0.429 & --    & --    \\
    LayoutXLM$_\text{base}$~\cite{xu2021layoutxlm}         & I+T+L & 0.794 & 0.548 & 0.521 & 0.423 \\
    LayoutXLM$_\text{large}$~\cite{xu2021layoutxlm}        & I+T+L & 0.823 & 0.640 & 0.581 & 0.536 \\
    \bottomrule
  \end{tabular}
\end{table}

\paragraph{Comparison with VLMs.}
\Cref{tab:vlm-comparison} presents the main results on FUNSD and XFUND.
Our 256M-parameter model outperforms all zero-shot VLMs on relation extraction and is competitive on entity recognition, while avoiding external OCR/layout systems.

\begin{table}[t]
  \caption{VLM comparison on FUNSD and XFUND. \cmark/\xmark\ indicates layout-aware (bounding box) evaluation. All baselines are evaluated zero-shot.}
  \label{tab:vlm-comparison}
  \centering
  \setlength{\tabcolsep}{5pt}
  \renewcommand{\arraystretch}{1.1}
  \begin{tabular}{lcc cccc}
    \toprule
    \multirow{2}{*}{\textbf{Model}} & \multirow{2}{*}{\textbf{Size}} & \multirow{2}{*}{\textbf{Loc.}} & \multicolumn{2}{c}{\textbf{FUNSD}} & \multicolumn{2}{c}{\textbf{XFUND}} \\
    \cmidrule(lr){4-5} \cmidrule(lr){6-7}
    & & & \textbf{SER} & \textbf{RE} & \textbf{SER} & \textbf{RE} \\
    \midrule
    Llama-3.2~\cite{grattafiori2024llama3}    & 11B & \xmark & 0.389 & 0.186 & --    & --    \\
    GPT-4o~\cite{openai2024gpt4}              & --  & \xmark & \textbf{0.554} & 0.274 & --    & --    \\
    Qwen2.5-VL~\cite{bai2025qwen25vl}         & 7B  & \cmark & 0.273 & 0.144 & 0.285 & 0.123 \\
    \midrule
    \rowcolor{hlrow}
    \textbf{Ours (SmolDocling)}                        & \textbf{256M} & \cmark & 0.514 & \textbf{0.275} & \textbf{0.412} & \textbf{0.206} \\
    \bottomrule
  \end{tabular}
\end{table}

\paragraph{Detailed comparison with Qwen2.5-VL.}
Since Qwen2.5-VL is the only baseline with box prediction, \Cref{tab:qwen-detailed} compares it to ours under the same layout-aware metric.
Our model is stronger on all public-set metrics; the private set is used only as supporting layout-diversity evidence.
It also runs at 0.27s/sample vs.\ 1.40s/sample on A100 GPUs ($>$5$\times$ faster).

\begin{table}[t]
  \caption{Layout-aware comparison with Qwen2.5-VL across all three datasets.}
  \label{tab:qwen-detailed}
  \centering
  \setlength{\tabcolsep}{4.5pt}
  \renewcommand{\arraystretch}{1.1}
  \begin{tabular}{lc cccccc}
    \toprule
    \multirow{2}{*}{\textbf{Model}} & \multirow{2}{*}{\textbf{Size}} & \multicolumn{2}{c}{\textbf{DocLayNetV2}} & \multicolumn{2}{c}{\textbf{FUNSD}} & \multicolumn{2}{c}{\textbf{XFUND}} \\
    \cmidrule(lr){3-4} \cmidrule(lr){5-6} \cmidrule(lr){7-8}
    & & \textbf{SER} & \textbf{RE} & \textbf{SER} & \textbf{RE} & \textbf{SER} & \textbf{RE} \\
    \midrule
    Qwen2.5-VL~\cite{bai2025qwen25vl}  & 7B  & 0.179 & 0.074 & 0.273 & 0.144 & 0.285 & 0.123 \\
    \rowcolor{hlrow}
    \textbf{Ours (SmolDocling)}                       & \textbf{256M} & \textbf{0.485} & \textbf{0.321} & \textbf{0.514} & \textbf{0.275} & \textbf{0.412} & \textbf{0.206} \\
    \bottomrule
  \end{tabular}
\end{table}

\paragraph{Metric sensitivity.}
All main layout-aware results use Levenshtein ratio $\rho_{\text{Lev}}\geq0.8$ and IoU$\geq0.7$.
With IoU threshold 0.0, removing the hard localization gate while keeping Hungarian assignment, our model increases from 0.514/0.275 to 0.679/0.376 on FUNSD and from 0.412/0.206 to 0.516/0.289 on XFUND (SER/RE); Qwen2.5-VL on XFUND increases from 0.285/0.123 to 0.359/0.175.
Thus relaxing localization raises absolute scores while preserving the trend; the stricter metric better reflects spatial grounding.

\subsection{Ablation Studies}
\label{sec:ablations}

We conduct ablations along two axes: the data pipeline (\cref{tab:ablation-data}) and the model configuration (\cref{tab:ablation-config}).
Unless stated otherwise, the main results in \cref{tab:vlm-comparison,tab:qwen-detailed} use the best overall configuration: \emph{all crops + aug.\ + synth.\ filled} data pipeline with \emph{link-in-key} strategy.

\paragraph{Data pipeline ablation.}
\Cref{tab:ablation-data} evaluates the contribution of each data pipeline component.
When no form-filling strategy is specified (first three rows), empty fields are excluded from training entirely.
Key findings are as follows:
(i)~\emph{Crop augmentation} is highly effective; using crops from all datasets vs.\ DocLayNetV2 crops only improves XFUND RE by ${\sim}$41\% (0.122$\to$0.172), confirming that focused key-value crops help the model learn entity semantics.
(ii)~\emph{Image transformations} (aug.) improve FUNSD performance (noisy scanned documents) while slightly decreasing DocLayNetV2 scores (clean PDFs), showing domain-specific effects.
(iii)~\emph{Synthetic form filling} consistently improves FUNSD and XFUND over both retaining unfilled fields as-is (unfilled) and excluding empty fields entirely, validating our data augmentation strategy. The synthetically filled variant (synth.\ filled) achieves the best overall balance, with the strongest RE on XFUND (0.206) and SER on FUNSD (0.543).

\begin{table}[t]
  \caption{Data pipeline ablation (layout-aware evaluation). DLNv2\,=\,DocLayNetV2.}
  \label{tab:ablation-data}
  \centering
  \setlength{\tabcolsep}{4pt}
  \renewcommand{\arraystretch}{1.1}
  \begin{tabular}{lcccccc}
    \toprule
    \multirow{2}{*}{\textbf{Variant}} & \multicolumn{2}{c}{\textbf{DLNv2}} & \multicolumn{2}{c}{\textbf{FUNSD}} & \multicolumn{2}{c}{\textbf{XFUND}} \\
    \cmidrule(lr){2-3} \cmidrule(lr){4-5} \cmidrule(lr){6-7}
    & \textbf{SER} & \textbf{RE} & \textbf{SER} & \textbf{RE} & \textbf{SER} & \textbf{RE} \\
    \midrule
    DLNv2 crops + aug.           & 0.401          & 0.287          & 0.479          & \underline{0.275} & 0.321          & 0.122          \\
    All crops                     & 0.425          & \underline{0.304} & 0.521          & 0.261          & 0.361          & 0.166          \\
    All crops + aug.              & 0.416          & 0.296          & 0.514          & \textbf{0.275} & 0.348          & \underline{0.172} \\
    All crops + aug. + unfilled         & \textbf{0.485} & \textbf{0.321} & \underline{0.523} & 0.239          & \underline{0.376} & 0.171          \\
    All crops + aug. + synth.\ filled$^\dagger$ & \underline{0.459} & 0.301          & \textbf{0.543} & 0.261          & \textbf{0.412} & \textbf{0.206} \\
    \bottomrule
  \end{tabular}
  \\[2pt]
  {\scriptsize $^\dagger$Configuration used for main results.}
\end{table}

\paragraph{Model configuration ablation.}
\Cref{tab:ablation-config} examines linking strategies and training configurations, all using the \emph{all crops + aug.\ + synth.\ filled} data pipeline.
Spatial ordering arranges DocTags elements by top-left coordinates, providing a consistent serialization that may conflict with logical reading order in complex layouts.

\emph{Spatial ordering} improves entity recognition (e.g., +1.4\% SER on FUNSD, +2.9\% on XFUND) but slightly decreases RE, likely because coordinate order does not always match logical reading order~\cite{wang2021layoutreader}.
\emph{Link-in-value} performs comparably to \emph{link-in-key} for SER but reduces RE, suggesting that anchoring links in keys helps before decoding the target value.
\emph{Weighted loss} (wt.\ loss, $\lambda{=}5$ on structural tokens) prioritizes graph markup amid many content tokens; it achieves the highest DLNv2 SER but decreases other metrics.
Since no single configuration dominates across all metrics, we select the \emph{link-in-key} configuration for the main results based on its consistently strong relation extraction performance.

\begin{table}[t]
  \caption{Model configuration ablation (layout-aware evaluation). DLNv2\,=\,DocLayNetV2. All variants use all crops + aug.\ + synth.\ filled.}
  \label{tab:ablation-config}
  \centering
  \setlength{\tabcolsep}{4pt}
  \renewcommand{\arraystretch}{1.1}
  \begin{tabular}{lcccccc}
    \toprule
    \multirow{2}{*}{\textbf{Config}} & \multicolumn{2}{c}{\textbf{DLNv2}} & \multicolumn{2}{c}{\textbf{FUNSD}} & \multicolumn{2}{c}{\textbf{XFUND}} \\
    \cmidrule(lr){2-3} \cmidrule(lr){4-5} \cmidrule(lr){6-7}
    & \textbf{SER} & \textbf{RE} & \textbf{SER} & \textbf{RE} & \textbf{SER} & \textbf{RE} \\
    \midrule
    Link-in-key$^\dagger$              & 0.459          & \textbf{0.301} & 0.543          & \textbf{0.261} & 0.412          & \textbf{0.206} \\
    Spatial + link-in-key              & 0.504          & \underline{0.282} & \textbf{0.557} & \underline{0.251} & 0.441          & \underline{0.194} \\
    Spatial + link-in-value            & \underline{0.506} & 0.267          & \underline{0.551} & 0.235          & \textbf{0.448} & 0.175          \\
    Spatial + link-in-value + wt.\ loss & \textbf{0.509} & 0.269          & 0.510          & 0.197          & \underline{0.444} & 0.178          \\
    \bottomrule
  \end{tabular}
  \\[2pt]
  {\scriptsize $^\dagger$Configuration used for main results.}
\end{table}

\subsection{Qualitative Analysis}
\label{sec:qualitative}

\begin{figure}[t]
  \centering
  \begin{subfigure}[b]{0.30\textwidth}
    \centering
    \begin{tcolorbox}[colframe=black, colback=white, arc=6pt, boxrule=1pt, left=0pt, right=0pt, top=0pt, bottom=0pt]
      \includegraphics[width=\textwidth]{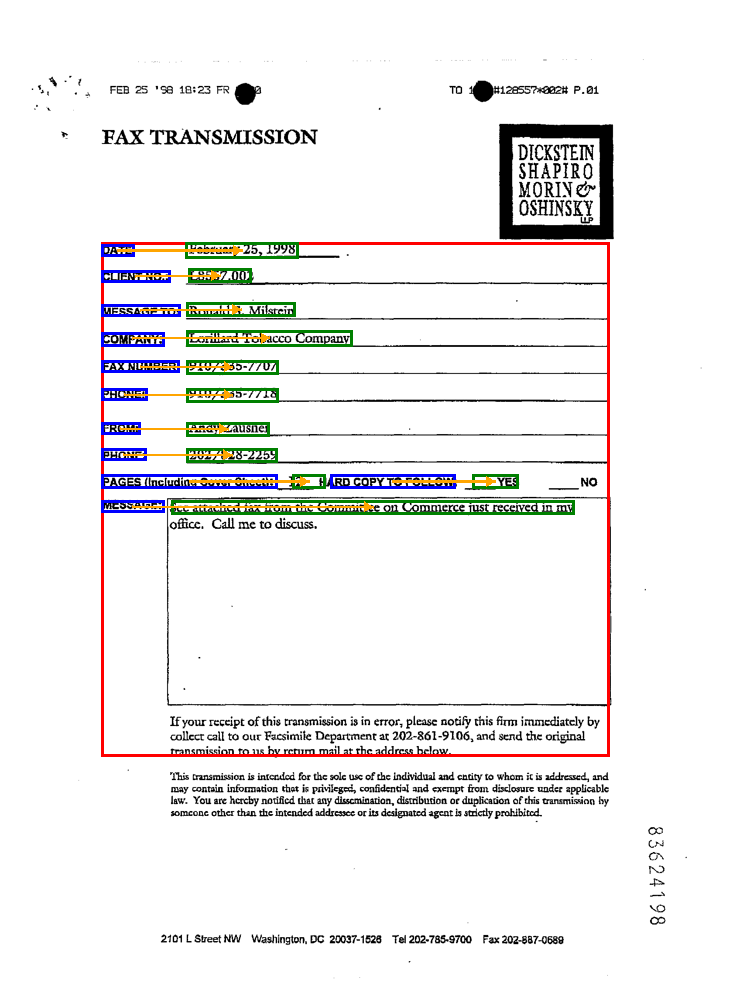}
    \end{tcolorbox}
    \caption{FUNSD}
    \label{fig:pred-funsd}
  \end{subfigure}
  \hfill
  \begin{subfigure}[b]{0.30\textwidth}
    \centering
    \begin{tcolorbox}[colframe=black, colback=white, arc=6pt, boxrule=1pt, left=0pt, right=0pt, top=0pt, bottom=0pt]
      \includegraphics[width=\textwidth]{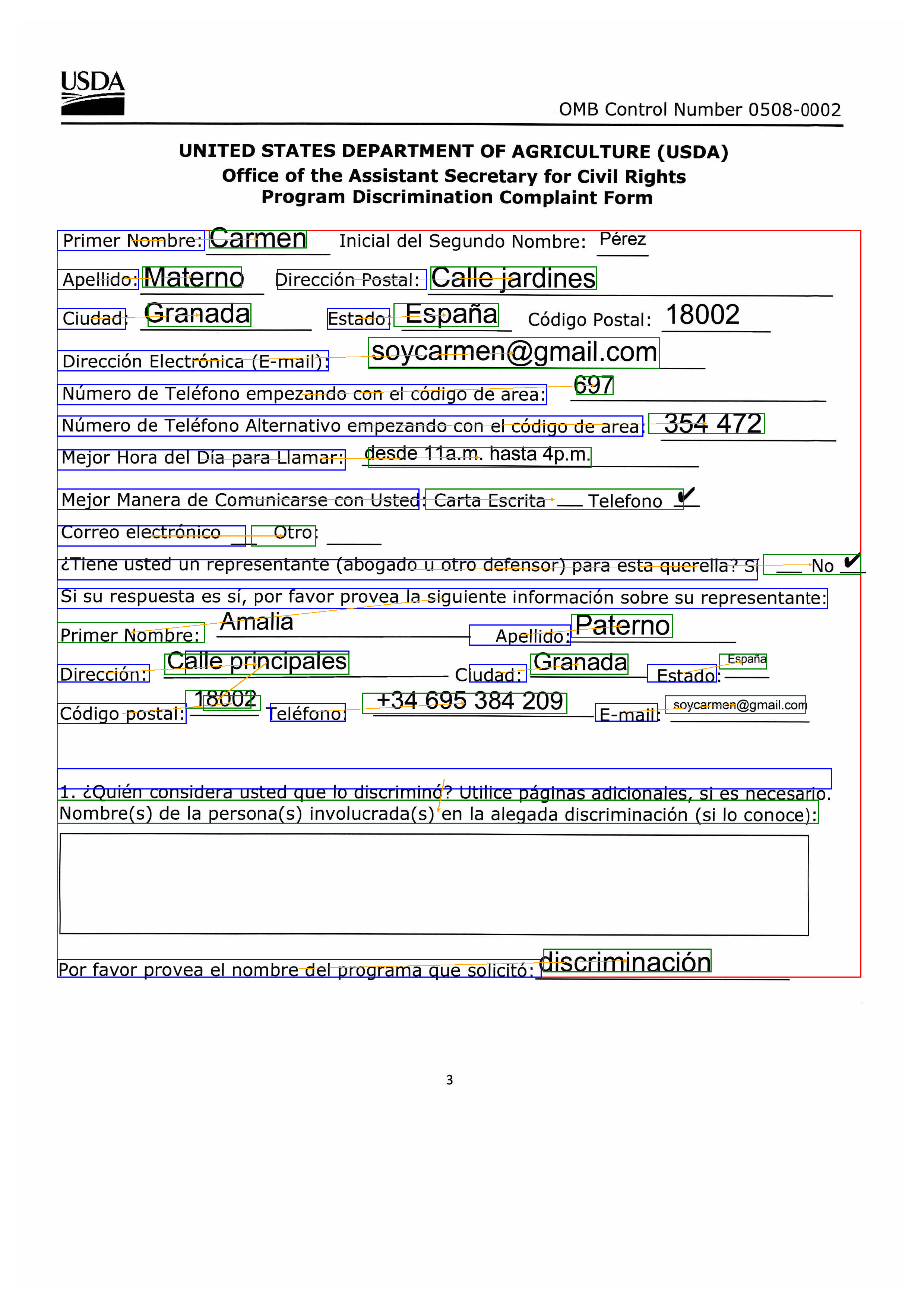}
    \end{tcolorbox}
    \caption{XFUND (Spanish)}
    \label{fig:pred-xfund-es}
  \end{subfigure}
  \hfill
  \begin{subfigure}[b]{0.30\textwidth}
    \centering
    \begin{tcolorbox}[colframe=black, colback=white, arc=6pt, boxrule=1pt, left=0pt, right=0pt, top=0pt, bottom=0pt]
      \includegraphics[width=\textwidth]{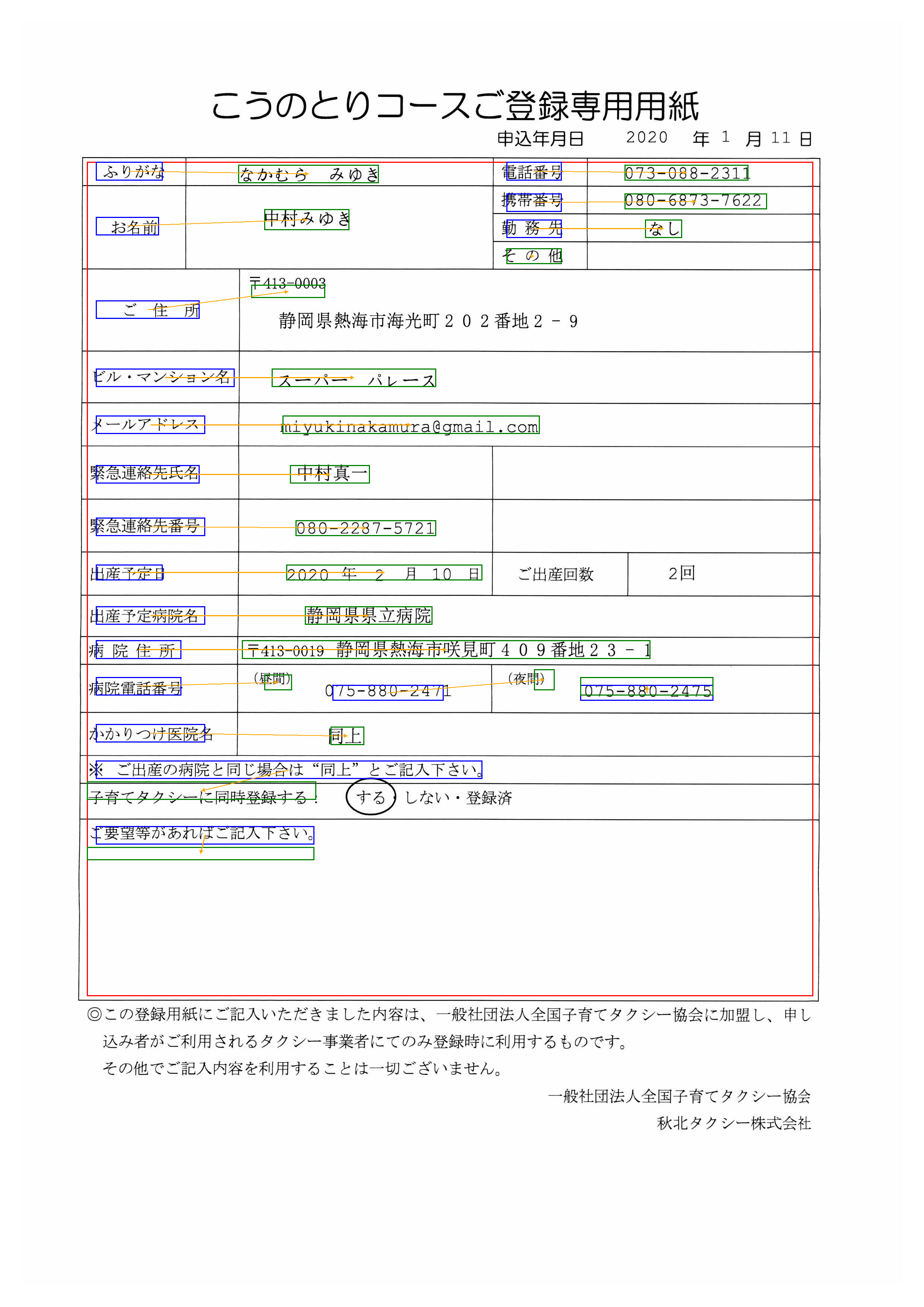}
    \end{tcolorbox}
    \caption{XFUND (Japanese)}
    \label{fig:pred-xfund-ja}
  \end{subfigure}
  \caption{Predictions on FUNSD and XFUND test documents. The model correctly identifies keys (blue), values (green), bounding boxes, and linking relationships (yellow arrows) across English, Spanish, and Japanese forms.}
  \label{fig:qualitative-funsd-xfund}
\end{figure}

\begin{figure}[t]
  \centering
  \begin{subfigure}[b]{0.32\textwidth}
    \centering
    \begin{tcolorbox}[colframe=black, colback=white, arc=6pt, boxrule=1pt, left=0pt, right=0pt, top=0pt, bottom=0pt]
      \includegraphics[width=\textwidth]{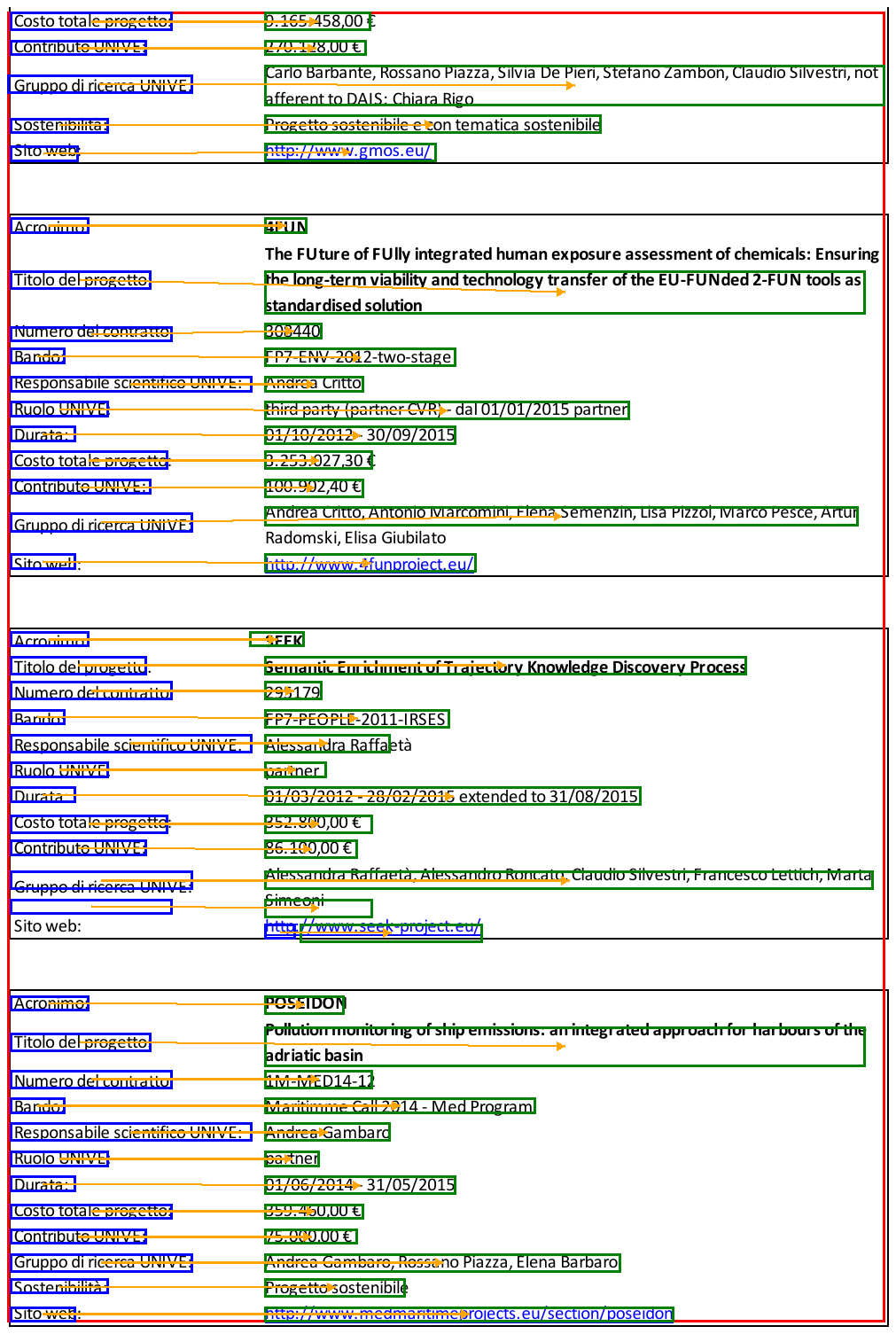}
    \end{tcolorbox}
    \caption{Financial document}
    \label{fig:pred-dlnv2-1}
  \end{subfigure}
  \hfill
  \begin{subfigure}[b]{0.32\textwidth}
    \centering
    \begin{tcolorbox}[colframe=black, colback=white, arc=6pt, boxrule=1pt, left=0pt, right=0pt, top=0pt, bottom=0pt]
      \includegraphics[width=\textwidth]{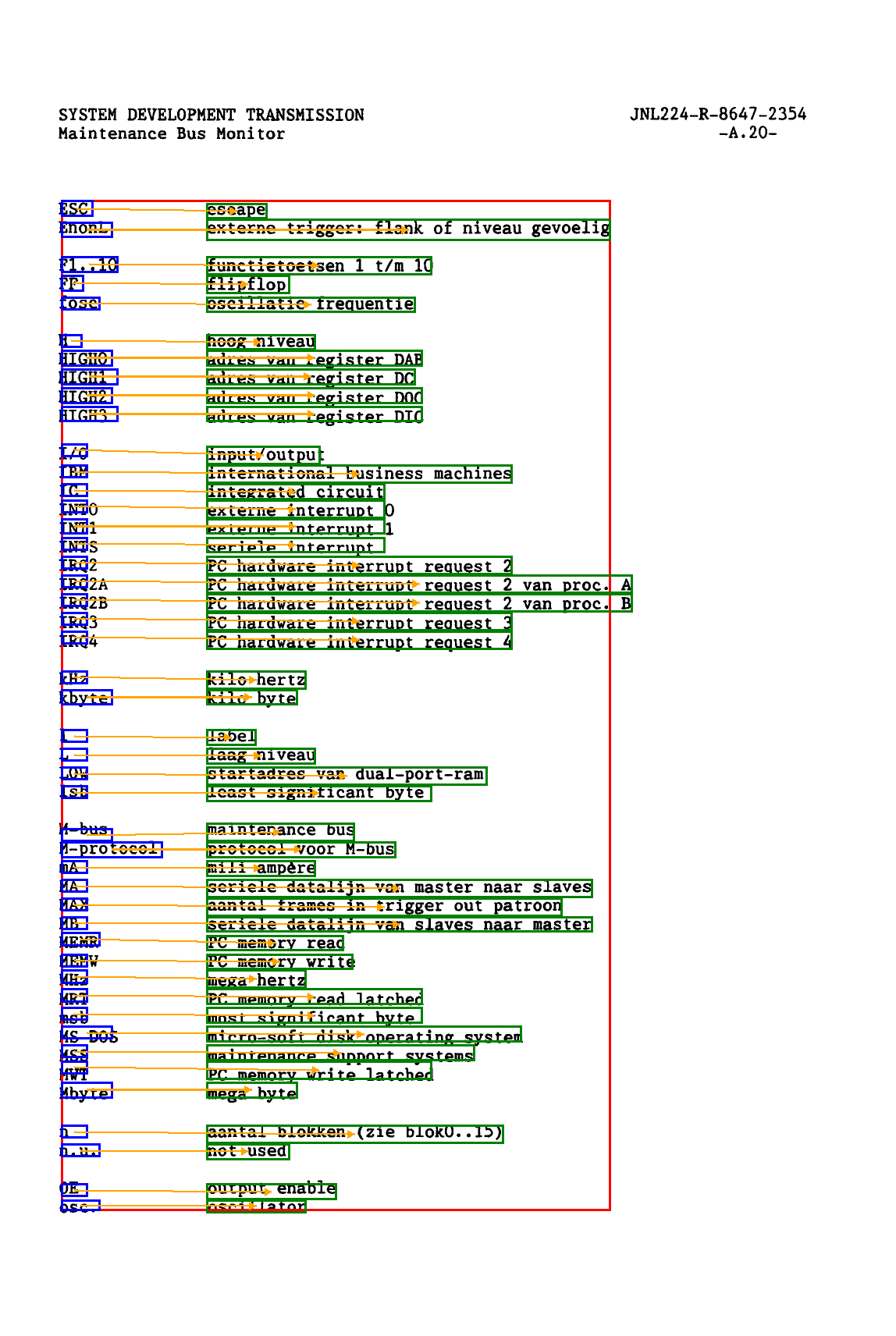}
    \end{tcolorbox}
    \caption{Form document}
    \label{fig:pred-dlnv2-3}
  \end{subfigure}
  \hfill
  \begin{subfigure}[b]{0.32\textwidth}
    \centering
    \begin{tcolorbox}[colframe=black, colback=white, arc=6pt, boxrule=1pt, left=0pt, right=0pt, top=0pt, bottom=0pt]
      \includegraphics[width=\textwidth]{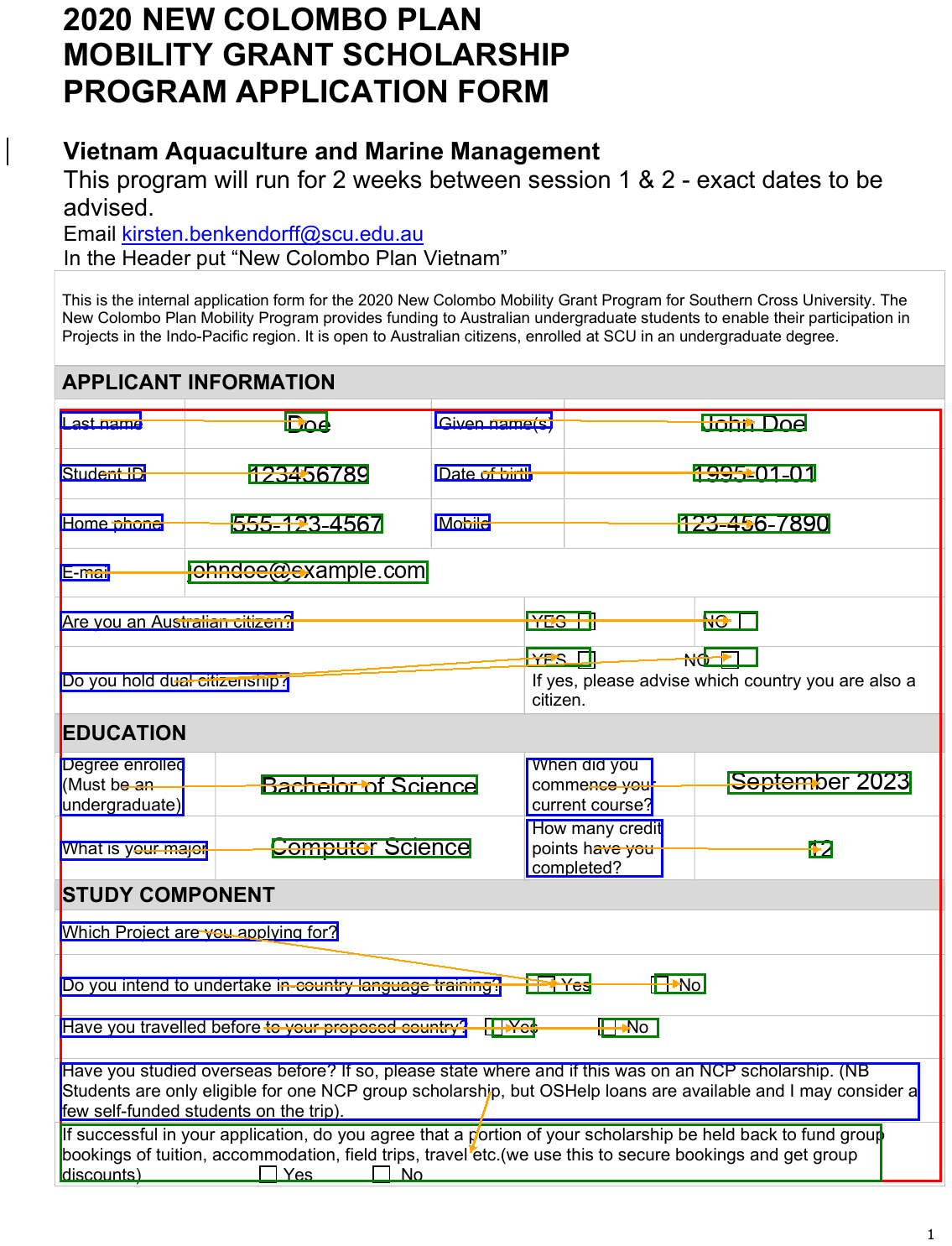}
    \end{tcolorbox}
    \caption{Report document}
    \label{fig:pred-dlnv2-4}
  \end{subfigure}
  \caption{Predictions on DocLayNetV2 test documents. The model generalizes across diverse document types: financial reports (a), structured forms (b), and report-style layouts (c).}
  \label{fig:qualitative-dlnv2}
\end{figure}

\Cref{fig:qualitative-funsd-xfund,fig:qualitative-dlnv2} show prediction examples across all three datasets.
In these visualizations, red boxes indicate key-value regions, blue boxes mark keys, green boxes mark values, and yellow arrows show predicted links.

The examples show noisy FUNSD scans, multilingual XFUND forms, and diverse DocLayNetV2 layouts including financial reports, dense forms, and report-style pages.

\paragraph{Error analysis.}
Examining failure cases across all three datasets reveals four recurring error patterns.
\emph{OCR errors} account for many false negatives, as misrecognized characters cause text matching to fail even when the entity is correctly classified and localized.
\emph{Missed links} occur when key-value pairs are spatially distant, repeated, or separated by other document content, suggesting the model relies partly on spatial proximity for linking.
\emph{Entity boundary errors} arise when multi-word keys or values are over- or under-segmented relative to the ground truth.
Finally, \emph{complex layouts} with deeply nested structures or many-to-many relationships spanning large distances remain the most challenging, consistent with the relatively low RE scores compared to SER across all experiments.

%% file: sections/conclusion.tex
\section{Conclusion}
\label{sec:conclusion}

This work demonstrates that a compact 256M-parameter vision-language model can perform end-to-end key-value extraction from document images, jointly solving identification, localization, and association in a single forward pass.
Two findings stand out from our experiments.
First, task-specific fine-tuning on a well-designed data pipeline matters more than model scale in the OCR-free setting: our 256M model outperforms zero-shot Qwen2.5-VL (7B) under layout-aware evaluation, despite being 27$\times$ smaller and over 5$\times$ faster at inference.
Second, the data pipeline components contribute unevenly: crop augmentation and synthetic form filling each provide large gains (up to 41\% on relation extraction), while choices like spatial ordering and link placement involve trade-offs between entity recognition and relation extraction with no single configuration dominating.
The approach is most suitable when deployment should operate directly on document images, avoid external OCR/layout components, and return spatially grounded key-value evidence for disambiguation or visual retrieval workflows.

\paragraph{Limitations.}
Our model does not yet match OCR/text-layout encoder models that receive pre-extracted text and boxes as input, though this comparison is inherently asymmetric: those models rely on stronger input assumptions while ours must perform OCR, classification, localization, and association from pixels.
The limited number of publicly available datasets with spatial key-value annotations restricts both training diversity and evaluation coverage.
Additionally, the lack of reading order information in current datasets forces coordinate-based serialization, which is suboptimal for complex multi-column layouts where spatial ordering diverges from logical reading flow~\cite{wang2021layoutreader}.

\paragraph{Future work.}
Scaling the architecture to 1B--7B parameters may close the gap with OCR-based methods, as capabilities like OCR accuracy, spatial reasoning, and multilingual understanding strongly correlate with model size.
Incorporating reading order prediction into the serialization pipeline could improve autoregressive generation consistency on complex layouts.
Finally, the model's compact size (256M parameters) makes it a candidate for on-device deployment~\cite{xu2024ondevice} in document processing workflows where latency and privacy constraints preclude cloud-based inference.

%% file: main.bbl
\begin{thebibliography}{10}
\providecommand{\url}[1]{\texttt{#1}}
\providecommand{\urlprefix}{URL }
\providecommand{\doi}[1]{https://doi.org/#1}

\bibitem{allal2025smollm2}
Allal, L.B., Lozhkov, A., Bakouch, E., Bl\'azquez, G.M., Penedo, G., Tunstall,
  L., Marafioti, A., Kydl\'i\v{c}ek, H., Lajar\'in, A.P., Srivastav, V.,
  Lochner, J., Fahlgren, C., Nguyen, X.S., Fourrier, C., Burtenshaw, B.,
  Larcher, H., Zhao, H., Zakka, C., Morlon, M., Raffel, C., von Werra, L.,
  Wolf, T.: Smollm2: When smol goes big -- data-centric training of a small
  language model (2025)

\bibitem{auer2024docling}
Auer, C., Lysak, M., Nassar, A., Dolfi, M., Livathinos, N., Vagenas, P., Ramis,
  C.B., Omenetti, M., Lindlbauer, F., Dinkla, K., Mishra, L., Kim, Y., Gupta,
  S., de~Lima, R.T., Weber, V., Morin, L., Meijer, I., Kuropiatnyk, V., Staar,
  P.W.J.: Docling technical report (2024)

\bibitem{bai2025qwen25vl}
Bai, S., Chen, K., Liu, X., Wang, J., Ge, W., Song, S., Dang, K., Wang, P.,
  Wang, S., Tang, J., Zhong, H., Zhu, Y., Yang, M., Li, Z., Wan, J., Wang, P.,
  Ding, W., Fu, Z., Xu, Y., Ye, J., Zhang, X., Xie, T., Cheng, Z., Zhang, H.,
  Yang, Z., Xu, H., Lin, J.: Qwen2.5-vl technical report (2025)

\bibitem{blecher2023nougat}
Blecher, L., Cucurull, G., Scialom, T., Stojnic, R.: Nougat: Neural optical
  understanding for academic documents (2023)

\bibitem{buslaev2020albumentations}
Buslaev, A., Iglovikov, V.I., Khvedchenya, E., Parinov, A., Druzhinin, M.,
  Kalinin, A.A.: Albumentations: Fast and flexible image augmentations.
  Information  \textbf{11}(2) (2020). \doi{10.3390/info11020125}

\bibitem{Cao_2023_graphRevisedIE}
Cao, P., Wu, J.: Graphrevisedie: Multimodal information extraction with
  graph-revised network. Pattern Recognition  \textbf{140},  109542 (2023).
  \doi{10.1016/j.patcog.2023.109542}

\bibitem{cheng2024xformparser}
Cheng, X., Zhang, H., Yang, J., Li, X., Zhou, W., Liu, F., Wu, K., Guan, X.,
  Sun, T., Wu, X., Li, T., Li, Z.: Xformparser: A simple and effective
  multimodal multilingual semi-structured form parser (2024)

\bibitem{chi2021infoxlm}
Chi, Z., Dong, L., Wei, F., Yang, N., Singhal, S., Wang, W., Song, X., Mao,
  X.L., Huang, H., Zhou, M.: Infoxlm: An information-theoretic framework for
  cross-lingual language model pre-training (2021)

\bibitem{conneau2020xlmroberta}
Conneau, A., Khandelwal, K., Goyal, N., Chaudhary, V., Wenzek, G., Guzm\'an,
  F., Grave, E., Ott, M., Zettlemoyer, L., Stoyanov, V.: Unsupervised
  cross-lingual representation learning at scale (2020)

\bibitem{devlin2019bert}
Devlin, J., Chang, M.W., Lee, K., Toutanova, K.: Bert: Pre-training of deep
  bidirectional transformers for language understanding (2019)

\bibitem{grattafiori2024llama3}
Grattafiori, A., et~al.: The llama 3 herd of models (2024)

\bibitem{hong2022bros}
Hong, T., Kim, D., Ji, M., Hwang, W., Nam, D., Park, S.: {BROS}: A pre-trained
  language model focusing on text and layout for better key information
  extraction from documents. In: Proceedings of the AAAI Conference on
  Artificial Intelligence. vol.~36, pp. 10767--10775 (2022),
  \url{https://ojs.aaai.org/index.php/AAAI/article/view/21322}

\bibitem{huang2022layoutlmv3}
Huang, Y., Lv, T., Cui, L., Lu, Y., Wei, F.: Layoutlmv3: Pre-training for
  document ai with unified text and image masking (2022)

\bibitem{jaumefunsd2019}
Jaume, G., Ekenel, H.K., Thiran, J.P.: Funsd: A dataset for form understanding
  in noisy scanned documents (2019)

\bibitem{kim2022donut}
Kim, G., Hong, T., Yim, M., Nam, J., Park, J., Yim, J., Hwang, W., Yun, S.,
  Han, D., Park, S.: Ocr-free document understanding transformer (2022)

\bibitem{king2019idcdata}
King, T.: 80 percent of your data will be unstructured in five years. Solutions
  Review  (2019)

\bibitem{kuhn1955hungarian}
Kuhn, H.W.: The hungarian method for the assignment problem. Naval Research
  Logistics Quarterly  \textbf{2}(1-2),  83--97 (1955).
  \doi{10.1002/nav.3800020109}

\bibitem{lee2022formnet}
Lee, C.Y., Li, C.L., Dozat, T., Perot, V., Su, G., Hua, N., Ainslie, J., Wang,
  R., Fujii, Y., Pfister, T.: Formnet: Structural encoding beyond sequential
  modeling in form document information extraction (2022)

\bibitem{levenshtein_1966}
Levenshtein, V.I.: Binary codes capable of correcting deletions, insertions and
  reversals. Soviet Physics Doklady  \textbf{10}(8) (1966)

\bibitem{li2022trocr}
Li, M., Lv, T., Chen, J., Cui, L., Lu, Y., Florencio, D., Zhang, C., Li, Z.,
  Wei, F.: Trocr: Transformer-based optical character recognition with
  pre-trained models (2022)

\bibitem{li2021structext}
Li, Y., Qian, Y., Yu, Y., Qin, X., Zhang, C., Liu, Y., Yao, K., Han, J., Liu,
  J., Ding, E.: Structext: Structured text understanding with multi-modal
  transformers (2021)

\bibitem{liu2019roberta}
Liu, Y., Ott, M., Goyal, N., Du, J., Joshi, M., Chen, D., Levy, O., Lewis, M.,
  Zettlemoyer, L., Stoyanov, V.: Roberta: A robustly optimized bert pretraining
  approach (2019), \url{https://arxiv.org/abs/1907.11692}

\bibitem{lu2025laytextllm}
Lu, J., Yu, H., Wang, Y., Ye, Y., Tang, J., Yang, Z., Wu, B., Liu, Q., Feng,
  H., Wang, H., Liu, H., Huang, C.: A bounding box is worth one token:
  Interleaving layout and text in a large language model for document
  understanding (2025)

\bibitem{luo2023geolayoutlm}
Luo, C., Cheng, C., Zheng, Q., Yao, C.: Geolayoutlm: Geometric pre-training for
  visual information extraction (2023)

\bibitem{lysak2023otsl}
Lysak, M., Nassar, A., Livathinos, N., Auer, C., Staar, P.: Optimized table
  tokenization for table structure recognition. In: Document Analysis and
  Recognition - ICDAR 2023: 17th International Conference, San Jos\'{e}, CA,
  USA, August 21–26, 2023, Proceedings, Part II. p. 37–50. Springer-Verlag,
  Berlin, Heidelberg (2023). \doi{10.1007/978-3-031-41679-8_3},
  \url{https://doi.org/10.1007/978-3-031-41679-8_3}

\bibitem{marafiotismolvlm2025}
Marafioti, A., Zohar, O., Farr\'e, M., Noyan, M., Bakouch, E., Cuenca, P.,
  Zakka, C., Allal, L.B., Lozhkov, A., Tazi, N., Srivastav, V., Lochner, J.,
  Larcher, H., Morlon, M., Tunstall, L., von Werra, L., Wolf, T.: Smolvlm:
  Redefining small and efficient multimodal models (2025)

\bibitem{nassarsmoldocling2025}
Nassar, A., Marafioti, A., Omenetti, M., Lysak, M., Livathinos, N., Auer, C.,
  Morin, L., de~Lima, R.T., Kim, Y., Gurbuz, A.S., Dolfi, M., Farr\'e, M.,
  Staar, P.W.J.: Smoldocling: An ultra-compact vision-language model for
  end-to-end multi-modal document conversion (2025)

\bibitem{openai2024gpt4}
OpenAI, et~al.: Gpt-4 technical report (2024)

\bibitem{Pfitzmann_2022_doclaynet}
Pfitzmann, B., Auer, C., Dolfi, M., Nassar, A.S., Staar, P.: Doclaynet: A large
  human-annotated dataset for document-layout segmentation. In: Proceedings of
  the 28th ACM SIGKDD Conference on Knowledge Discovery and Data Mining. pp.
  3743--3751 (2022). \doi{10.1145/3534678.3539043}

\bibitem{smith2007tesseract}
Smith, R.: An overview of the tesseract ocr engine. In: Ninth International
  Conference on Document Analysis and Recognition (ICDAR 2007). vol.~2, pp.
  629--633 (2007). \doi{10.1109/ICDAR.2007.4376991}

\bibitem{vaswani2023attentionisallyouneed}
Vaswani, A., Shazeer, N., Parmar, N., Uszkoreit, J., Jones, L., Gomez, A.N.,
  Kaiser, L., Polosukhin, I.: Attention is all you need (2023)

\bibitem{wang2023docllm}
Wang, D., Raman, N., Sibue, M., Ma, Z., Babkin, P., Kaur, S., Pei, Y.,
  Nourbakhsh, A., Liu, X.: Docllm: A layout-aware generative language model for
  multimodal document understanding (2023)

\bibitem{wang2022lilt}
Wang, J., Jin, L., Ding, K.: Lilt: A simple yet effective language-independent
  layout transformer for structured document understanding (2022)

\bibitem{wang2021layoutreader}
Wang, Z., Xu, Y., Cui, L., Shang, J., Wei, F.: Layoutreader: Pre-training of
  text and layout for reading order detection (2021)

\bibitem{xu2024ondevice}
Xu, J., Li, Z., Chen, W., Wang, Q., Gao, X., Cai, Q., Ling, Z.: On-device
  language models: A comprehensive review (2024)

\bibitem{xu2022layoutlmv2}
Xu, Y., Xu, Y., Lv, T., Cui, L., Wei, F., Wang, G., Lu, Y., Florencio, D.,
  Zhang, C., Che, W., Zhang, M., Zhou, L.: Layoutlmv2: Multi-modal pre-training
  for visually-rich document understanding (2022)

\bibitem{Xu_2020layoutlm}
Xu, Y., Li, M., Cui, L., Huang, S., Wei, F., Zhou, M.: Layoutlm: Pre-training
  of text and layout for document image understanding. In: Proceedings of the
  26th ACM SIGKDD International Conference on Knowledge Discovery; Data Mining.
  pp. 1192--1200 (2020). \doi{10.1145/3394486.3403172}

\bibitem{xu2021layoutxlm}
Xu, Y., Lv, T., Cui, L., Wang, G., Lu, Y., Florencio, D., Zhang, C., Wei, F.:
  Layoutxlm: Multimodal pre-training for multilingual visually-rich document
  understanding (2021)

\bibitem{xu-2022-xfund}
Xu, Y., Lv, T., Cui, L., Wang, G., Lu, Y., Florencio, D., Zhang, C., Wei, F.:
  {XFUND}: A benchmark dataset for multilingual visually rich form
  understanding. In: Findings of the Association for Computational Linguistics:
  ACL 2022. pp. 3214--3224 (2022). \doi{10.18653/v1/2022.findings-acl.253}

\bibitem{zhaisiglip2023}
Zhai, X., Mustafa, B., Kolesnikov, A., Beyer, L.: Sigmoid loss for language
  image pre-training (2023)

\end{thebibliography}
